%% file: main.tex
\documentclass[runningheads]{llncs}

\usepackage{graphicx}
\usepackage{hyperref}

\hypersetup{
  colorlinks   = true, 
  urlcolor     = blue, 
  linkcolor    = blue, 
  citecolor   = blue 
}

\usepackage{enumitem,kantlipsum}
\usepackage{amsmath,amssymb,amsfonts}
\usepackage{textcomp}
\usepackage{multirow}
\usepackage{booktabs}
\usepackage{afterpage}
\usepackage{paralist}

\usepackage{times}

\usepackage{amsmath}
\usepackage{algorithm}
\usepackage[noend]{algpseudocode}
\newcommand{\mycomment}[1]{} 

\usepackage{array}
\newcolumntype{L}[1]{>{\raggedright\let\newline\\\arraybackslash\hspace{0pt}}m{#1}}
\newcolumntype{C}[1]{>{\centering\let\newline\\\arraybackslash\hspace{0pt}}m{#1}}
\newcolumntype{R}[1]{>{\raggedleft\let\newline\\\arraybackslash\hspace{0pt}}m{#1}}

\newcommand{\upquote}{\text{\textquotesingle}}
\newcommand\tab[1][0.2cm]{\hspace*{#1}}

\begin{document}

\title{Expressive Reasoning Graph Store: A Unified Framework for Managing RDF and Property Graph Databases}

\titlerunning{Expressive Reasoning Graph Store}

\author{Sumit Neelam, Udit Sharma, Sumit Bhatia, Hima Karanam, Ankita Likhyani, Ibrahim Abdelaziz, Achille Fokoue, L.V. Subramaniam}
\authorrunning{S. Neelam et al.}
\institute{IBM Research AI\\
}

\maketitle              

\begin{abstract}
Resource Description Framework (RDF) and Property Graph (PG) are the two most commonly used data models for representing, storing, and querying graph data. We present Expressive Reasoning Graph Store (ERGS) -- a graph store built on top of JanusGraph (a Property Graph store) that also allows storing and querying of RDF datasets. First, we describe how RDF data can be translated into a Property Graph representation and then describe a query translation module that converts SPARQL queries into a series of Gremlin traversals. The converters and translators thus developed can allow any Apache Tinkerpop compliant graph database to store and query RDF datasets. We demonstrate the effectiveness of our proposed approach using JanusGraph as the base Property Graph store and compare its performance with standard RDF systems.

\textbf{Resource Link:} \url{https://github.com/IBM/expressive-reasoning-graph-store}\\
\textbf{License:} Apache-2.0
\end{abstract}

\input{files/intro}
\input{files/SystemArchitecture.tex}
\input{files/SchemaTranslation.tex}
\input{files/QueryEngine.tex}
\input{files/Experiments.tex}
\input{files/conclusions.tex}

\footnotesize
\bibliographystyle{splncs04}
\bibliography{references}

\end{document}

%% file: files/intro.tex
\section{Introduction}
Resource Description Framework (RDF)~\cite{rdf} and Property Graph (PG)~\cite{property-graph} are the two most prominent data models for representing graph data. 
While both the data models allow for intuitive modeling of graph data, there is poor interoperability between the two models making it hard to adopt systems designed for one model to support the other data model.

We consider the task of \textit{storing and querying RDF graphs on a Property Graph store}. There are many situations that warrant such a capability. There are certain operations such as computing PageRank~\cite{pagerank}, finding paths~\cite{paths} and graph traversals for computing knowledge graph embeddings~\cite{embeddings} that are well suited for a Property Graph architecture and can benefit from the Property Graph systems optimized for graph traversals~\cite{pg-shortcomings}. Further, RDF has been adopted by the World Wide Consortium (W3C) as the standard for describing resources in the Semantic Web and there exist a number of standard RDF datasets in various important domains such as finance, healthcare, and life sciences. Moreover, in many enterprise and production scenarios, the costs of maintaining multiple graph databases may prove to be prohibitive.

\noindent
\textbf{Related Work:} 
Hartig~\cite{Hartig14} proposed a formalization of the Property Graph data model and presented a formal framework for translating between the RDF and PG data models. Along similar lines, Tomaszuk~\cite{yars} and Matsumoto et al.~\cite{rdfInPG} also propose different transformations to store RDF data into Property Graph stores. However, all these methods solve only piece of the puzzle in that the proposed transformations allow storage of RDF datasets in Property Graph stores, however, the transformed data can then only be queried by means of the query language of the underlying graph database (such as Gremlin) and SPARQL queries over the original dataset are not supported. Thakkar et al.~\cite{gremlinator2018} on the other hand, have described Gremlinator, a query translator to convert SPARQL queries into Gremlin traversals for executing graph pattern matching queries over Property Graph stores.

We build upon and extend the formal framework by Hartig~\cite{Hartig14} for transforming RDF to Property Graphs and develop a query translator to convert SPARQL queries into their equivalent Gremlin traversals. These transformations allow us to treat any Property Graph system as a RDF store and we could perform any standard RDF operation over Property Graph stores. We present a detailed comparison with these two works later in the paper when we discuss our proposed transformations. Hartig~\cite{Hartig14} also proposed a new data model called RDF* which is meant to close the gap between RDF and Property Graph data models. Some native RDF stores started adopting this new model to benefit from its ability to annotate triples with metadata information. However, to the best of our knowledge, the gap still exists as graph databases are still scattered between the two data models. Through our proposed system, we aim to have a unified framework where users can store, query and reason with both data models efficiently. 

\noindent
\textbf{Our Contributions:} To the best of our knowledge, we present the first complete solution to enable storage and querying of RDF datasets in Property Graph stores. To achieve this, we first present our approach to transform RDF datasets into Property Graph model (Section~\ref{sec:rdf2pg}). We then describe our query translation module (Section~\ref{sec:queryEngine}) that converts a given SPARQL query into a Gremlin traversal. We implement our proposed data and query translation modules on top of JanusGraph and use LUBM, BSBM, and WatDiv benchmarks to compare the performance of resulting  \emph{JanusGraph based RDF Store} with competent RDF engines Virtuoso and Blazegraph (Section~\ref{sec:experiments}). Finally, we discuss the limitations of our proposed transformations and offer interesting directions for further work in bridging the gap between RDF and Property Graph systems (Section~\ref{sec:conclusions}). Our developed system is open-sourced under a permissive Apache license and is  available at \url{https://github.com/IBM/expressive-reasoning-graph-store}.

%% file: files/SystemArchitecture.tex
\section{System Architecture}
\begin{figure}
	\centering
	\includegraphics[width=1\columnwidth]{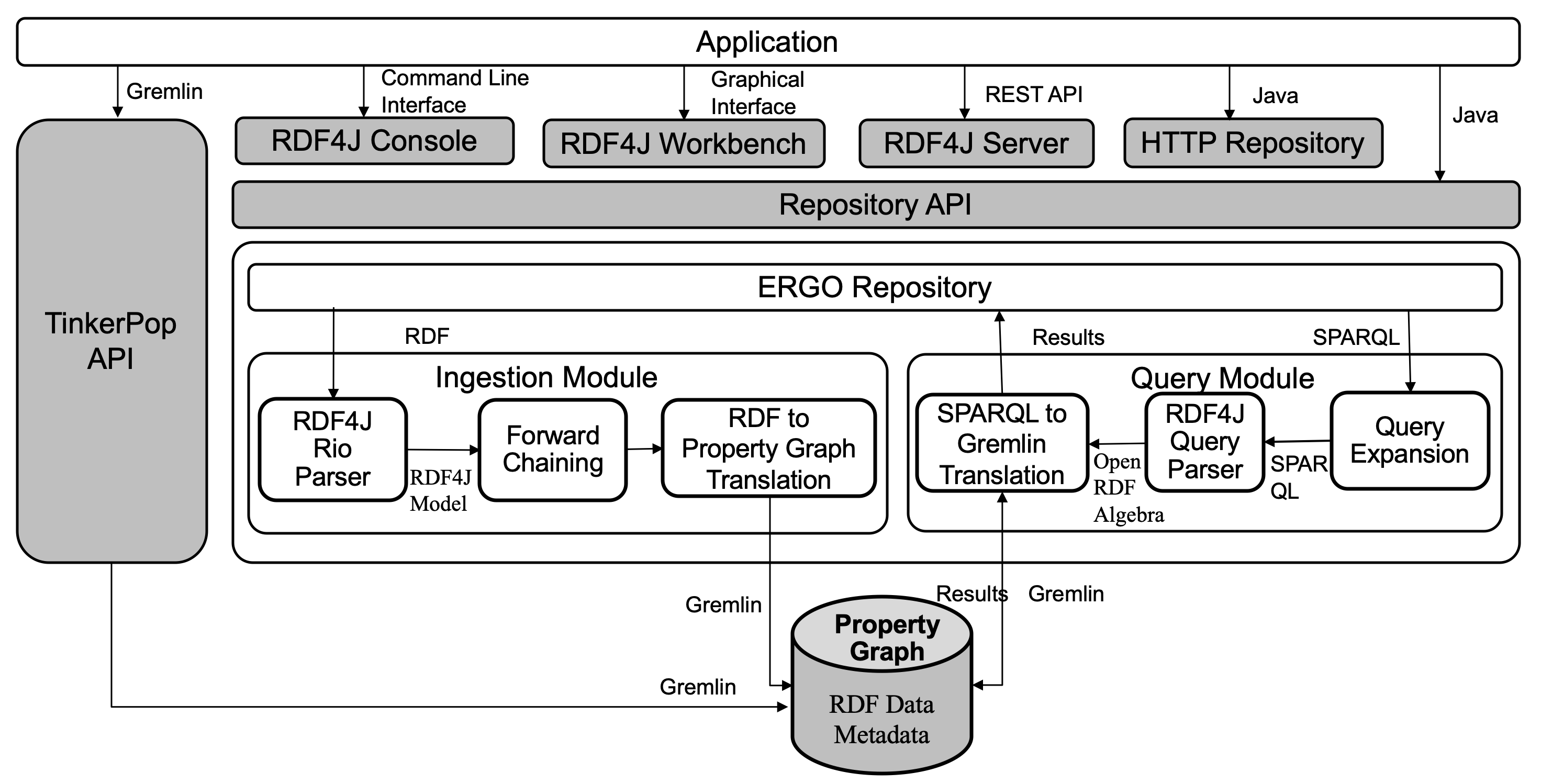}
	\caption{System Architecture}
	\label{fig:SysArch}
\end{figure}

Figure~\ref{fig:SysArch} presents the high-level architecture of our proposed system. We implemented the RDF4J\footnote{\url{https://rdf4j.org}} repository interface for creating a new datastore on top of an underlying Apache Tinkerpop\footnote{\url{http://tinkerpop.apache.org}} compliant Property Graph store along with all the required interfaces to support ingestion and querying of RDF data. We now describe the major components of our proposed system. 

\noindent
{\bf Ingestion Module:} implements the core RDF4J repository APIs that enable handling and ingesting the input data through various RDF4J supported input mechanisms. In the current implementation we support the RDF data model and a subset of OWL ontology constructs that can be associated with a given RDF graph. Section~\ref{sec:rdf2pg} provides more details about our approach of mapping RDF data to Property Graph framework and schema creation.

\noindent
{\bf Storage Layer:} We chose JanusGraph\footnote{\url{https://janusgraph.org}} (a Property Graph store) as our underlying Property Graph system to store the translated RDF graphs. We have used JanusGraph with Hbase\footnote{\url{https://hbase.apache.org}} as its underlying backend storage and utilize various indexing mechanisms provided by JanusGraph to achieve better SPARQL query performance (Section~\ref{sec:rdf2pg}). Along with the RDF data, various additional metadata are also stored in the JanusGraph as additional nodes that are utilized by the query processing layer for effective translation of SPARQL queries to Gremlin traversals.

    
 \noindent
 {\bf Query Processing Module:} handles the input SPARQL queries or calls to RDF4J exploration APIs. We utilize the metadata stored during the ingestion phase along with the input SPARQL query to create appropriate Gremlin traversals to produce the desired results. There are three main components in the query processing module: \begin{inparaenum}[i)] \item a query expansion component that  generates an expanded query to support RDFS (and limited OWL) reasoning to generate inferred results; \item a SPARQL query parser that parses a given SPARQL and generates the corresponding parse tree; and \item a SPARQL to Gremlin translation that produces a series of Gremlin traversals to execute the SPARQL query.
 \end{inparaenum}Section~\ref{sec:queryEngine} describes in detail the process of SPARQL to Gremlin translation.

 \noindent
 {\bf API Layer:} This layer is the user-facing layer which hosts all the APIs in the form of REST and JavaAPIs for a user to interact with the underlying system. It currently has all the RDF4J ingestion, querying and exploration APIs. RDF4J provides four different types of interaction using various modes.
 \begin{inparaenum}[i)]
\item RDF4J console: a command-line application that can be used for accessing and modifying the data. \item RDF4J Workbench: provides different graphical interfaces for interacting with the underlying store using SPARQL and other endpoints. \item RDF4J Server: these are a set of REST APIs which allow the user to interact with the RDF store using HTTP protocol. \item HTTP repository: it provides a proxy for remote RDF4J repository, such that users could use Java APIs to interact with the repository just like a local repository. 
\end{inparaenum}

\noindent
{\bf Docker Image:} The system can be realized using Docker based container provided along with the source code. The container provides portable and flexible distribution of the system without need to compile the source code itself. The source code and the detailed documentation for the proposed system is also available under the Apache License. 
 

    
 \noindent

%% file: files/SchemaTranslation.tex
\section{Mapping RDF Data Model to Property Graph}
\label{sec:rdf2pg}

We have used Property Graph model provided by Apache Tinkerpop and 
defined it formally using modified definition of Rodriguez~\cite{Rodriguez}. 
We define the property graph as $G = (V, E, \rho, \lambda, \sigma)$ where $V$ is the set of vertices and $E$ is the set of edges such that $E \subseteq (V \times V)$. Here, $\rho: (V \cup E) \to L$ is a complete function that maps vertices and edges to string label $L$. $\lambda : ((V \cup E) \times \Sigma^{\star}) \to (U\backslash \ (V \cup E))$ is a partial multivalued function that maps a pair of an element (node or edge) and string ($\Sigma^{\star}$) to universal set $U$ (contains the set of all property values) excluding vertices and edges, and $\sigma: ((V \cup E) \times \Sigma^{\star} \times (U\backslash \ (V \cup E))) \to (U\backslash \ (V \cup E))$ is a partial function that maps element's property key and value to universal set U excluding V and E, i.e., defines meta-properties.


RDF data is generally serialized in a triple format. As a result of the data transformation process, the input triple data is converted into the Property Graph model described above. In addition, we also compute and store metadata as subgraph that contains information about how each predicate is represented in the resultant Property Graph. A predicate can either be mapped as a node property or as an edge to subject node of the triple. As we will see later, metadata is useful in SPARQL query translation and it allows users to directly execute SPARQL queries, just like an RDF store, on the underlying Property Graph structure. As specified in the W3C standards, there can be three types of objects in an RDF triple: \textit{primitives}, \textit{IRIs} and \textit{blank nodes}. IRIs and primitives together are called resources or entities. The resource denoted by an IRI is called its \textit{Referent}, and the resource denoted by a primitive value is called its \textit{Literal}. Literals have datatypes that define the range of possible values. Special kinds of literals (e.g., language-tagged strings) denote plain-text strings in a natural language.




\subsection{Basic Schema Translation}
\begin{figure}[t!]
	\centering
	\includegraphics[scale=0.27]{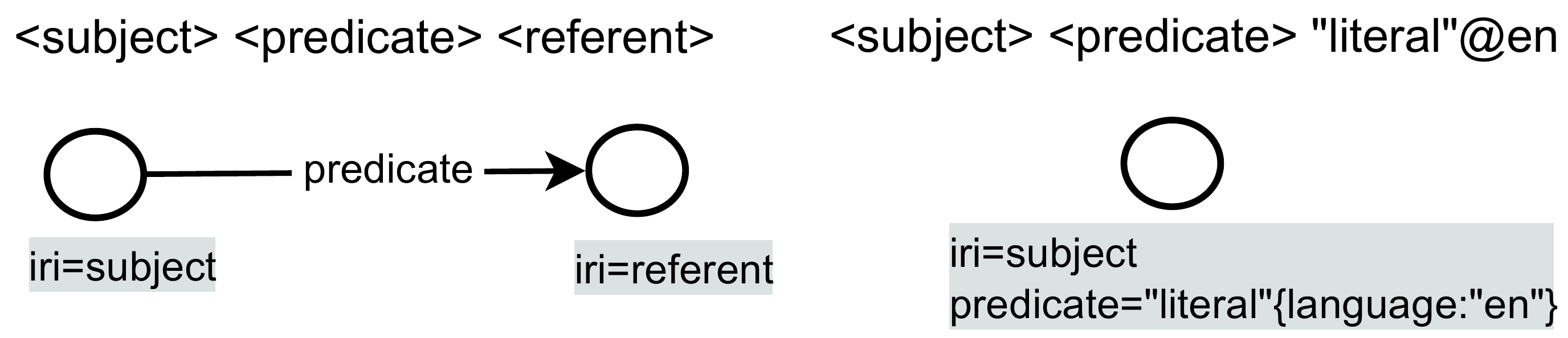}
	\caption{Triples Transform Strategy}
	\label{fig:tripleStrategy}
\end{figure}

Schema translation, as discussed by Hartig~\cite{Hartig14}, is a space-efficient and optimized data transformation method that utilizes the object type of the triples to reduce the number of nodes and edges in the resultant graph. Our proposed mapping builds upon this idea for schema transformation and incorporates mappings to enable the handling of metadata to aid in the process of query translation.  In the simplest form of transformation,  each RDF triple \textit{t}\{\textit{s,p,o}\} can be uniformly converted into two nodes, one each for \textit{s} and \textit{o} with \textit{IRI} as property, and an edge with label \textit{p}. During transformation, we use this simple mapping for triples with a referent or a blank node. The left side of Figure~\ref{fig:tripleStrategy} presents the case for this type of mapping. In contrast, triples that have literals as objects can be modeled differently for reducing the size of the graph. Literal objects are of primitive type and these can be seen as sink nodes in the native mapping. Hence, literal objects are stored as properties of the subject nodes for the corresponding triples (as shown in the right part of Figure~\ref{fig:tripleStrategy}). Besides reducing the size of the graph, this approach also helps in improving query execution times as less number of nodes lead to a reduced number of traversals required. During transforming \textit{Blank Node}, graph nodes is created and a blank node id is assigned during storage. Other associated information of the literal (such as datatype and language information) are stored as meta-property, i.e, properties of the property.

\subsection{Metadata for Supporting Query Translation} 
Metadata is stored as subgraph in same graph, but it is hidden from user. The metadata captures how each predicate present in the input RDF data is represented in the transformed Property Graph. This information is primarily useful for the query execution module for efficiently planning the graph traversals. Note that in a gremlin query, different types of gremlin operations (or traversals) are used to fetch node properties or adjacent nodes. Therefore, when a triple is required to be fetched for a specific predicate in SPARQL, the query engine module utilizes the information stored in the metadata to select the appropriate operation to fetch the triple.

Structurally, the metadata subgraph contains one node for each distinct predicate value in the RDF input and stores the type of mapping used. The node in the metadata can have three possible values (stored as property in the Property Graph store): \textit{literal, referent} and \textit{mixed}. If all the triples containing a given predicate have objects of type literal, then the type of predicate metanode is set as \textit{literal}. Likewise, the type of a predicate metanode is set to \textit{referent} when all the triples associated with the predicate have IRI objects. For predicates where the associated triples can have both an IRI or a literal as object type, the type of the predicate metanode is set as \textit{mixed}. We also collect some data statistics and store as metadata for improving query performance. 

\subsection{Parallel Ingestion and Index Utilization}

Graph loading can be time consuming operation for large datasets. Therefore, for making loading process faster we have developed parallel approach also. Tinkerpop based graph can become inconsistent with duplicate vertices for random parallel load. Therefore vertex creation is done keeping control on isolation. 
Hence, input data is read in batches and each batch is loaded in two cycles. In first cycle, each batch is divided into multiple node disjoint partitions. For each partition its nodes and intra-partition edges are independently loaded in parallel. This ensures that no more then one thread access same graph element and no duplicate vertex is created. In second cycle, inter-partition edges are loaded.  

Property Graph databases also provide various indexing mechanisms to make graph traversals more efficient. We make use of the following different indexing mechanisms provided by JanusGraph for faster query execution.
\begin{inparaenum}[i)]
    \item The IRI property of all the nodes are indexed. It allows us to search for the nodes with the given IRI. 
    \item All the data properties are indexed. It allows us to do exact/regex match for any data property. 
    \item We also index incident edges label of each node which allows us to search for the nodes with given edge label.
\end{inparaenum}
Note that this is a JanusGraph specific optimization. Our approach is general and can be implemented on top of any Tinkerpop compliant graph store, and appropriate indexing mechanisms of the corresponding graph store can be utilized.

 
\begin{figure}[ht!]
	\centering
	\includegraphics[width=1\columnwidth]{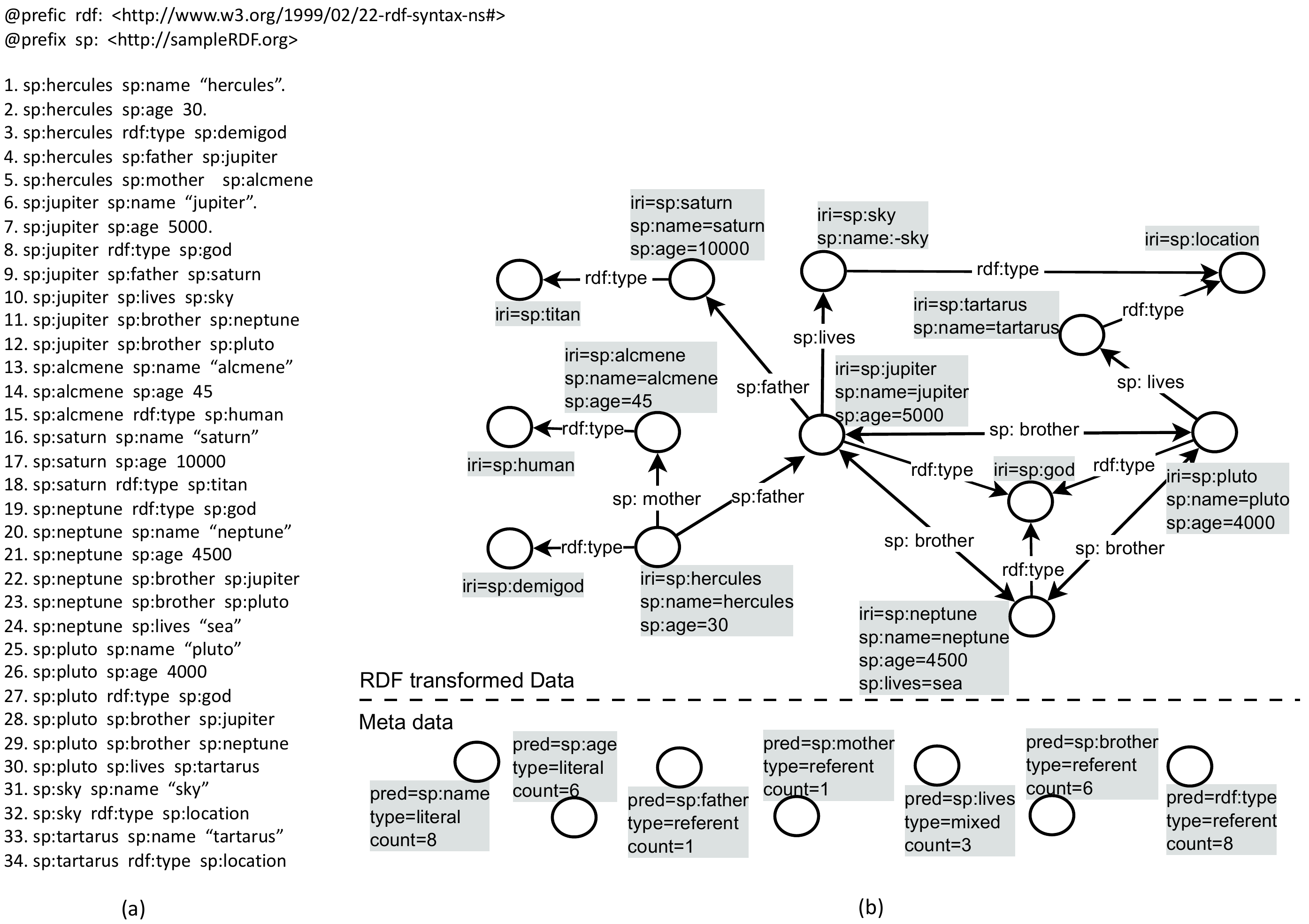}
	\caption{Sample RDF data and Transformed Property Graph of Roman Mythology}
	\label{fig:inputrdf}
\end{figure}


\subsection{Illustration of Data Transformation Process}
We now explain the RDF to Property Graph translation process with the help of an example. Consider the example RDF graph in Figure~\ref{fig:inputrdf}(a) representing characters from Roman mythology. This graph has been taken from sample graphs as provided by JanusGraph. The resultant Property Graph representation as obtained by applying the proposed transformation is presented in Figure~\ref{fig:inputrdf}(b). Consider the triples for subject \textit{sp:hercules}. For the first two triples in Figure~\ref{fig:inputrdf}(a),   the predicates are \textit{sp:name} and \textit{sp:age} and the objects for both are of type literal. Therefore, we store these two predicates as vertex properties. On the other hande,  the other two triples for \textit{sp:hercules} with predicates \textit{sp:father} and \textit{sp:mother}, have referent objects, i.e., the object has IRI. These predicates are therefore represented as edges from \textit{sp:hercules} to \textit{sp:jupiter} and \textit{sp:alcmene}, respectively.  

Figure~\ref{fig:inputrdf}(b) also shows the metadata created as a result of transforming sample RDF data. A dotted line separates graph and its metadata subgraph for better readability. Input RDF has seven types of predicates which in effect creates seven meta nodes. Triples with predicate \textit{sp:name} have literal only and predicate \textit{sp:mother} contains IRI objects only, therefore their types are set as \textit{literal} and \textit{referent} respectively. Predicate \textit{sp:lives} meta node is \textit{mixed} type as triples with predicate holds both types of object (refer triples 10 and 24 in Figure~\ref{fig:inputrdf}(a)). 

\subsection{Ingestion with Forward Chaining}
For executing reasoning queries more efficiently current ingestion pipeline supports enabling forward chaining during data ingestion. Forward chaining process precomputes closure of data by taking TBox axioms into account. This facilitates to execute reasoning queries directly on data without doing complex computation. As part of ingestion, closures are computed incrementally and stored along with input data. Currently, we support RDFS reasoning and few OWL constructs which includes \textit{owl:inverseOf, owl:symmetric, owl:TransitiveProperty}. Currently, we do not have full support for forward chaining in parallel loading, e.g., we do not support parallel ingestion for forward chaining for \textit{owl:TransitiveProperty} because it can update other sub-graphs also. In future, we will add this feature with other OWL axioms.

%% file: files/QueryEngine.tex
\section{Query Engine: Graph Traversals for SPARQL Queries}
\label{sec:queryEngine}

We now describe the query translation process that allows us to support SPARQL querying over RDF data stored in an underlying Property Graph database (JanusGraph in our implementation).
SPARQL supports a wide variety of clauses that allow the users to perform different operations on the RDF graph. It is therefore required to transform different constructs present in the SPARQL 1.0 and SPARQL 1.1 to their corresponding gremlin traversal step. Table \ref{tab:sparqlStructure} lists the basic structure of a SPARQL 1.1 query and the constructs for which we need to develop equivalent Gremlin traversals.

\begin{table}[t]

    \begin{small}
    \centering
	\caption{Structure of a SPARQL query}
	\label{tab:sparqlStructure}
	
	\resizebox{\textwidth}{!}{%
    \begin{tabular}{@{}L{7cm} L{5cm}@{}}
    \toprule
        \textbf{Clauses} & \textbf{Description}\\ 
        \midrule
        \begin{tabular}[c]{@{}l@{}}
            PREFIX prefix: \textless{} iri ${\_}$ ref \textgreater\\ 
            BASE \textless{} iri ${\_}$ ref \textgreater
        \end{tabular}
        & 
        \textbf{Prefix Declaration} (optional)\\ 
        \midrule
        \noalign{\smallskip}
        
        \begin{tabular}[c]{@{}l@{}}
            SELECT  (DISTINCT$|$REDUCED)?\\
            (Var$|$Expression AS Var) $|$ \\ 
            DESCRIBE Var $|$\\   
            CONSTRUCT BGP $|$ \\ 
            ASK
        \end{tabular}
        &
        \begin{tabular}[c]{@{}l@{} }
            \textbf{Result Clause}\\
            var = \{?v\}$^+|$ *\\
            TP = s p o. \\
            BGP = \{TP$^+$\}
        \end{tabular}\\
        
        \midrule
        \noalign{\smallskip}
        
        FROM (DefaultGraph$|$NamedGraph) 
        & 
       \textbf{Dataset definition} (optional) \\ 
        
       \midrule
       \noalign{\smallskip}
        
        \begin{tabular}[c]{@{}l@{}}
            WHERE \{\\ 
            BGP UNION$|$OPTIONAL$|$MINUS BGP\\
            FILTER Condition\\
            SERVICE\\
            BIND\\
            \}
        \end{tabular}
        &
        \begin{tabular}[c]{@{}l@{}}
            \textbf{Query Pattern}\\
            BGP = \{TP$^+$\}\{PP$^+$\}\{VC$^+$\}\\
            TP = s p o.\\
            PP = s PathExpression o.\\
            VC = VALUES DataBlock
        \end{tabular} \\ 
        
        \midrule
        \noalign{\smallskip}
        
        \begin{tabular}[c]{@{}l@{}}
            GROUP BY ...\\
            HAVING\\
            ORDER BY (ASC$|$DESC)? (Var$|$Expression)\\
            LIMIT INTEGER OFFSET INTEGER
        \end{tabular} 
        & 
        \textbf{Query Modifiers} (optional) \\ 
        
        \bottomrule
    \end{tabular}
    }
    \end{small}
\end{table}

\subsection{Comparison with Existing Work }
Gremlinator~\cite{gremlinator2018} is a plugin for Apache Tinkerpop that provides a compiler to transform SPARQL queries to their corresponding Gremlin traversals. While it allows the user to query Property Graphs using SPARQL, the underlying data is still viewed as a Property Graph and the end-user is expected to have the complete knowledge about how the RDF data is represented in the form of Property Graph. In addition, it also suffers from the following limitations: 
\begin{inparaenum}[i)]
    \item it assumes input SPARQL query has a prefix to distinguish between label-access traversal, out-edge traversal, property traversals, and property-value traversal. Such kind of information is not available in most of the standard queries. Moreover, such kind of prefixes prevents us from using the standard prefixes that are part of query; 
    \item The predicate p in a triple pattern \{s p o .\} can not be a variable; 
    \item	REGEX in filters of a graph pattern is not supported;
    \item	SPARQL Union query with unbalanced patterns is not supported;
    \item	in-edge traversals are not supported. Due to this certain categories of queries can not be handled by this plugin; and
    \item SPARQL 1.1 clauses are not supported. 
\end{inparaenum}

\subsection{Query Expansion and Preprocessing}
As a first step of the query translation process, we  expand the input SPARQL query as a union of multiple conjunctive queries to find some of the implicit solutions to the query. Typical approaches~\cite{QuOnto2005,dolby2009} uses TBox axioms to expand subset of DL constructs present in the query. Query expansion is guaranteed to find all the solutions, when the underlying knowledge base follows DL-lite logic. We have used an open-source library Quetzal\footnote{\url{https://github.com/Quetzal-RDF/quetzal}} for query expansion that can handle any SPARQL 1.0 query and OWL2-QL knowledge bases. 

The resulting conjunctive SPARQL queries are declarative in nature. This implies that the user can enter the triple patterns in any order without worrying about the order in which these patterns will be used during query execution. Gremlin supports both declarative and imperative queries. Gremlin declarative queries are formulated using the Match step\footnote{\url{http://tinkerpop.apache.org/docs/current/reference/\#match-step}} which consists of one or more traversal patterns similar to triples in a SPARQL query. So, SPARQL triples can be directly transformed to Gremlin patterns using the rule-based mapping~\cite{gremlinator2018}. However, there are some limitations associated with the Match Step --  
\begin{inparaenum}[i)]
\item all the match()-traversals must have a single start label;
\item certain kind of match patterns are not supported; and
\item very limited support for utilizing indexes.
\end{inparaenum} This prevents us from handling some of the declarative SPARQL queries and nested union queries. As part of query preprocessing, we reorder the query patterns in SPARQL declarative queries such that it can be directly translated using the SPARQL-Gremlin translator. This reordering is achieved by the following series of actions.
\begin{inparaenum}[i)]
     \item Select a triple pattern that utilizes indexes present on the graph. If there is no such Triple pattern select some random Triple Pattern. Add the triple pattern to the output query. Add subject and object to the visited list.
     \item Recursively add all the triple patterns where the subject or object is already visited.
     \item If some Triple pattern is still not part of the output query go to Step i. 
 \end{inparaenum}
 
 
\subsection{SPARQL to Gremlin Translation}
The proposed SPARQL to Gremlin translation module acts as a middle-ware between the query endpoint and the underlying Property Graph database such that the target system is viewed as an RDF store by the end-user. We use the metadata collected during the data ingestion phase (Section~\ref{sec:rdf2pg}) to differentiate between edge traversals and property traversals at the time of query translation. This allows us to avoid the use of non-standard prefixes employed by Gremlinator.

Table~\ref{tab:sampleQuery} presents a sample input SPARQL query and the resultant output Gremlin query generated by our query translation module for the example RDF and Property Graph data in Figure \ref{fig:inputrdf}. The table also provides the Gremlinator query that can be used to query the graph directly using the TinkerPop plugin. The abbreviated IRIs are used in the table for both Gremlin and Gremlinator to make the query more readable. The query finds all the nodes of type \textit{sp:god} and their father node if it exists. The gremlin traversal is initialized using g.inject(1).constant(\upquote N/A\upquote).as(\upquote v1\upquote, \upquote v2\upquote, ..., \upquote vn\upquote).  It marks all the query output variables as unbound. The variable can get bound during the remaining part of the query. If the variable remains unbound till the end of the query, then ``N/A"  is returned as the result for such variables. The traversal starts by fetching the node having  \textit{sp:god} as IRI and finds  all the incoming adjacent nodes connected with \textit{rdf:type} label, the IRI of these nodes are stored as label P, the traversal then optionally fetches the outgoing adjacent nodes with \textit{sp:father} label and stores the IRI in path label F. The output of the sample query is shown in Table \ref{tab:sampleOutput}. We now describe our approach to translate different SPARQL 1.1 constructs. We assume the reader to have a basic familiarity with gremlin traversals and direct the interested reader to Tinkerpop's gremlin manual for further details.

\begin{table}[t]
    \begin{small}
        \centering
        \caption{Sample SPARQL Query and its Gremlin and Gremlinator translations}
        \label{tab:sampleQuery}
        \begin{tabular}{@{}L{3.7cm}@{~~} L{4.7cm} @{~~} L{3.6cm}@{}}
            \toprule
            \textbf{SPARQL Input} & \textbf{Gremlin Output} & \textbf{Gremlinator}\\
            \midrule
             PREFIX ub: \textless{}http://sampleRDF.org\textgreater \newline 
                SELECT * WHERE \{\newline 
                \tab ?P rdf:type sp:god.\newline
                \tab OPTIONAL\{\newline
                \tab \tab ?P sp:father ?F.\newline
                \tab \}\newline
                \}         
            & 
            g.V().inject(1).constant(\upquote N/A\upquote).\newline
            as(\upquote P\upquote,\upquote F\upquote).V().has(\upquote iri\upquote,\upquote sp:god\upquote).\newline
            as(\upquote PTypev\upquote).inE(\upquote rdf:type\upquote).\newline
            outV().as(\upquote Pv\upquote).properties(\upquote iri\upquote).\newline
            value().as(\upquote P\upquote).optional(\newline
            \_\_.select(\upquote Pv\upquote).outE(\upquote sp:father\upquote).\newline
            inV().as(\upquote Fv\upquote).properties(\upquote iri\upquote).\newline
            value().as(\upquote F\upquote).select(\upquote P\upquote,\upquote F\upquote)
            &
            SELECT * WHERE\{\newline 
                \tab ?Pv e:type ?PTypev.\newline
                \tab ?PTypev v:iri "sp:god".\newline
                \tab ?Pv v:iri ?P.\newline
                \tab OPTIONAL\{\newline
                \tab \tab ?Pv e:father ?Fv.\newline
                \tab \tab ?Fv v:iri ?F.\newline
                \tab \}\newline
                \} 
            \\ 
            \bottomrule
        \end{tabular}
    \end{small}
\end{table}
\begin{table}[t]
    \centering
    \begin{small}
    \caption{Sample Query Output}
    \label{tab:sampleOutput}
    \begin{tabular}{@{}c @{~~} c @{}}
        \toprule
        \textbf{P} & \textbf{F} \\
        \midrule
         \textless http://sampleRDF.org/jupiter\textgreater       
        & 
         \textless http://sampleRDF.org/saturn\textgreater 
        \\ 
        
        \textless http://sampleRDF.org/pluto\textgreater  & 
        N/A
        \\ 
        
        \textless http://sampleRDF.org/neptune\textgreater       
        & 
        N/A
        \\ 
        \bottomrule
    \end{tabular}
    \end{small}
\end{table}

\begin{algorithm}[t]
\begin{small}
    \caption{Translation of Triple Pattern}\label{TP}
    \begin{algorithmic}[1]
      \Procedure{translateTriplePattern}{$T,s,p,o$}
            \If {$\textit{\underline{type}}(p)=literal$} 
               \State $translateLiteralTriplePattern(T,s,p,o)$
            \ElsIf {$\textit{\underline{type}}(p)=referent$} 
                \State $translateReferentTriplePattern(T,s,p,o)$
            \Else 
                \State $T \gets \underline{append}(T,union(translateReferentTriplePattern(T,s,p,o),\newline translateLiteralTriplePattern(T,s,p,o)))$
            \EndIf
        \EndProcedure
    \end{algorithmic}
\end{small}
\end{algorithm}
\begin{algorithm}[tbh]
\begin{small}
    \caption{Translation of Triple Pattern with Literal object}\label{TPWithDP}
    \begin{algorithmic}[1]
      \Procedure{translateLiteralTriplePattern}{$T,s,p,o$}
            \If {$\textit{\underline{isVariable}}(s)$} 
                \State $T \gets \underline{append}(T,select(\upquote s_v\upquote).values(\upquote iri\upquote).as(\upquote s\upquote)) $
            \Else 
                \State $T \gets \underline{append}(T,select(\upquote s_v\upquote).has(\upquote iri\upquote, \upquote s\upquote))$
            \EndIf
            \If {$\textit{\underline{isVariable}}(p)$} 
                \State $T \gets \underline{append}(T,select(\upquote s_v\upquote).properties().as(\upquote p_v\upquote).key().as(\upquote p\upquote)) $
            \Else 
                \State $T \gets \underline{append}(T,select(\upquote s_v\upquote).properties(\upquote p\upquote).as(\upquote p_v\upquote))$
            \EndIf
            \If {$\textit{\underline{isVariable}}(o)$} 
                \State $T \gets \underline{append}(T,select(\upquote p_v\upquote).value().as(\upquote o\upquote))$
            \Else 
                \State $T \gets \underline{append}(T,select(\upquote p_v\upquote).value().is(\upquote o\upquote))$
            \EndIf
            
        \EndProcedure
    \end{algorithmic}
    \end{small}
\end{algorithm}
\begin{algorithm}[tbh]
\begin{small}
    \caption{Translation of Triple Pattern with IRI object}\label{TPWithOP}
    \begin{algorithmic}[1]
      \Procedure{translateReferentTriplePattern}{$T,s,p,o$}
            \If {$\textit{\underline{isVariable}}(s)$} 
                \State $T \gets \underline{append}(T,select(\upquote s_v\upquote).values(\upquote iri\upquote).as(\upquote s\upquote)) $
            \Else 
                \State $T \gets \underline{append}(T,select(\upquote s_v\upquote).has(\upquote iri\upquote, \upquote s\upquote))$
            \EndIf
            \If {$\textit{\underline{isVariable}}(p)$} 
                \State $T \gets \underline{append}(T,select(\upquote s_v\upquote).outE().as(\upquote p_v\upquote).label().as(\upquote p\upquote)) $
            \Else 
                \State $T \gets \underline{append}(T,select(\upquote s_v\upquote).outE(\upquote p\upquote).as(\upquote p_v\upquote))$
            \EndIf
            \If {$\textit{\underline{isVariable}}(o)$} 
                \State $T \gets \underline{append}(T,select(\upquote p_v\upquote).inV().as(\upquote o_v\upquote).values(\upquote iri\upquote).as(\upquote o\upquote))$
            \Else 
                \State $T \gets \underline{append}(T,select(\upquote p_v\upquote).inV().as(\upquote o_v\upquote).has(\upquote iri\upquote, \upquote o\upquote))$
            \EndIf
        \EndProcedure
    \end{algorithmic}
\end{small}
\end{algorithm}

\noindent
\begin{inparaenum}[1)]
    \item \textbf{Translation of prefix declaration}: Prefixes are handled by expanding any prefixes present in the query with full IRI and then applying any further operations on the resulting query. 
    
    \noindent
    \item \textbf{Translation of Query Pattern}: SPARQL query pattern consists of one or more Basic Graph Pattern connected with OPTIONAL/UNION/MINUS where each BGP consists of one of more Triple Pattern, Property Path  or Values Clause.
    
    \noindent
    \begin{inparaenum}[a)]
        \item \textit{Translation of Triple Pattern}:  We maintain a list of query variables that are already visited by the current traversal. If the subject of the triple pattern is already visited, Algorithm \ref{TP} is used for the translation. It uses Algorithm \ref{TPWithOP}, Algorithm \ref{TPWithDP} or both depending upon whether the type of predicate is \textit{literal}, \textit{referent}, or a \textit{variable}, respectively. The procedure \textit{isVariable()} returns true if the term is a variable and false otherwise and the procedure \textit{append()} concatenates two traversals together.  If the object of the triple pattern is already visited, we use procedure similar to Algorithm \ref{TPWithOP} with direction of edge-traversal reversed. If neither of subject and object is visited by the current traversal, add a V() step followed by the has() step (if possible to utilize indexes), before following Algorithm \ref{TP}. The variables that could get bound to literals during the query execution are not added to the list of visited query variables. It allows us to handle the cases of object-object joins in which the objects are literals. Also note that blank nodes in the query is treated just like another variable during the translation process.

        \noindent
        \item \textit{Translation of Property Path}: SPARQL 1.1 supports PredicatePath, InversePath, SequencePath, AlternativePath, ZeroOrMorePath, OneOrMorePath, ZeroOrOnePath and NegatedPropertySet as property path expression. PredicatePath, InversePath, SequencePath, AlternativePath and NegatedPropertySet are handeled by RDF4J SPARQL parser by converting it into equivalent SPARQL 1.0 query. ZeroOrMorePath is handled using emit().repeat(out(\upquote predicate\upquote) /(in(\upquote predicate\upquote)) whereas OneOrMorePath is handled using repeat(out(\upquote predicate \upquote)/(in(\upquote predicate \upquote)). emit() steps along with steps in Algorithm \ref{TPWithOP}.
        
        \noindent
        \item \textit{Translation of Values Clause}: VALUES clause is handled by converting it into union() step for each of the bindings in the Data Block.
        
        \noindent
       \item \textit{Translation of Basic Graph Pattern}: A Basic Graph Pattern is translated by translating each of the Triple patterns present in the query in order.  Let BGP=\{TP\textsubscript{1}, TP\textsubscript{2}, TP\textsubscript{3}, ..., TP\textsubscript{n}\} and $\phi$(TP\textsubscript{i}) be the gremlin translation for Traversal Pattern TP\textsubscript{i} . Then the translation of BGP is $\phi$(BGP)= $\phi$(TP\textsubscript{1}).$\phi$(TP\textsubscript{2}). $\phi$(TP\textsubscript{3}).  ...  $\phi$(TP\textsubscript{n})
       
       \noindent
       \item \textit{Translation of BGP connectors, Filters and Bind}: Table \ref{tab:filterTranslation} contains the rules to translate connectors, filters and bind operations. The table describes translation for only a subset of FILTER expressions and BIND expressions, but similar approach can be used for handling other operators like   \textless,  \textless=,  \textgreater, \textgreater=, !=, STRENDS and  CONTAINS. Here we would like to mention that textRegex() step is not yet supported by TinkerPop, but is provided by some of the graph providers like JanusGraph.
    \end{inparaenum}
    
    \noindent
    \item \textbf{Translation of Query Modifiers}: Table \ref{tab:queryModifiersTranslation} describes the translation of SPARQL query modifiers to equivalent Gremlin Traversal step. The table describes the translation of only COUNT aggregate, but similar approach can be used for handling other aggregates like SUM, MIN, MAX, AVG, and SAMPLE. 
    
    \noindent
    \item \textbf{Translation of Result Clause}:  Table \ref{tab:resultClauseTranslation} explains the translation of SPARQL result clause where different query types like SELECT, CONSTRUCT, ASK and DESCRIBE can be converted to equivalent Gremlin traversal step. 
  
\end{inparaenum}
\begin{table}[t]
            \centering
        	\caption{Translation of Connectors, Filters and Bind}
        	\label{tab:filterTranslation}
        	\begin{small}
        	\resizebox{\textwidth}{!}{%
            \begin{tabular}{@{}L{5.3cm} @{~~~} L{6.8cm}@{}}
                \toprule
                \textbf{Operation OP} & \textbf{Gremlin Traversal GT=$\phi$(OP)} \\ 
                \midrule
                BGP FILTER (?X = value) & $\phi$(BGP).select(\upquote X\upquote).where(is(eq(\upquote value\upquote)))\\ 
                \noalign{\smallskip}
                BGP FILTER (?X = ?Y) & $\phi$(BGP).select(\upquote X\upquote).where(eq(\upquote Y\upquote))\\ 
                \noalign{\smallskip}
                BGP FILTER (?X IN list)  & $\phi$(BGP).select(\upquote X\upquote).where(is(within(\upquote list\upquote)))\\ 
                \noalign{\smallskip}
                 BGP FILTER (STRSTARTS(?X,v))  & $\phi$(BGP).select(\upquote X\upquote).where(is(startingWith(\upquote v\upquote)))\\ 
                \noalign{\smallskip}
                BGP FILTER regex(?X, value)  & $\phi$(BGP).select(\upquote X\upquote).where(is(\textbf{textRegex}(\upquote v\upquote)))\\
                \noalign{\smallskip}
                BGP FILTER langMatches(?X,"EN")  & $\phi$(BGP).select(\upquote Xv\upquote).values(\upquote language\upquote).\newline where(is(eq(\upquote EN\upquote)))\\ 
                \noalign{\smallskip}
                BGP  FILTER (bound(?X))  & $\phi$(BGP).select(\upquote X\upquote).where(is(neq(\upquote N/A\upquote)))\\ 
                \noalign{\smallskip}
                BGP FILTER (cond1 $||$ cond2)  & $\phi$(BGP).or($\phi$(cond1),$\phi$(cond2))\\ 
                \noalign{\smallskip}
                BGP FILTER (cond1 $\&\&$ cond2)  & $\phi$(BGP).and($\phi$(cond1),$\phi$(cond2))\\ 
                \noalign{\smallskip}
                BGP FILTER (!cond)  & $\phi$(BGP).not($\phi$(cond))\\ 
                \noalign{\smallskip}
                BGP1 FILTER EXISTS BGP2  & $\phi$(BGP1).V().where($\phi$(BGP1))\\  
                \noalign{\smallskip}
                BGP1 FILTER NOT EXISTS BGP2 & $\phi$(BGP1).V().where(not($\phi$(BGP1)))\\ 
                \noalign{\smallskip}
                BGP1 UNION BGP2     & union($\phi$(BGP1),$\phi$(BGP2)) \\ 
                \noalign{\smallskip}
                BGP1 OPTIONAL BGP2 & $\phi$(BGP1).optional($\phi$(BGP2))\\ 
                \noalign{\smallskip}
                 BGP1 MINUS BGP2 & $\phi$(BGP1).where(not($\phi$(BGP1)))\\ 
                \noalign{\smallskip}
                 BGP BIND(?died - ?born AS ?age) & $\phi$(BGP).math(\upquote died - born\upquote).as(\upquote age\upquote)\\ 
                \noalign{\smallskip}
                BGP BIND(FLOOR(?X) AS ?Y) & $\phi$(BGP).math(\upquote floor  X\upquote).as(\upquote Y\upquote)\\ 
                \bottomrule
            \end{tabular}
            }
            \end{small}
        \end{table}
        
\begin{table}[ht]
        \centering
    	\caption{Translation of Query Modifiers}
    	\label{tab:queryModifiersTranslation}
        \begin{small}
        \resizebox{\textwidth}{!}{%
        \begin{tabular}{@{}L{4cm} @{~~~} L{8cm}@{}}
            \toprule
            \textbf{Query Modifier QM} & \textbf{Gremlin Traversal GT=$\phi$(QM)} \\ 
            \midrule
            ORDER BY (ASC $|$ DESC) ?(Var $|$ Exp)      & $\phi$(QP).order().by(‘var’$|\phi$(Exp),asc$|$desc) \\ 
            \noalign{\smallskip}
            QP LIMIT n                          & $\phi$(QP).range(0,n)\\ 
            \noalign{\smallskip}
            QP OFFSET n                         & $\phi$(QP).range(n,-1)\\ 
            \noalign{\smallskip}
            QP LIMIT n OFFSET m                 & $\phi$(QP).range(m,m+n)\\ 
            \noalign{\smallskip}
            (DISTINCT $|$ REDUCED)                  & $\phi$(QP).dedup()\\ 
            \noalign{\smallskip}
            SELECT varList, (COUNT(?var) AS ?count) WHERE\{ QP\} GROUP BY ?g1 ?g2 … HAVING condition & $\phi$(QP).group().by(select(\upquote g1\upquote,\upquote g2\upquote,...)).by(fold().
            match(\_\_.as(\upquote group\upquote).unfold().select(\upquote var\upquote).
            count().as(\upquote count\upquote).select(\upquote count\upquote)).unfold().
            match( \_\_.as(\upquote group\upquote).select(keys).select(\upquote g1\upquote).as(\upquote g1\upquote), \_\_.as(\upquote group\upquote).select(keys).select(\upquote g2\upquote).as(\upquote g2\upquote),
            ... \newline
            \_\_.as(\upquote group\upquote).select(values).select(\upquote count\upquote) .as(\upquote count\upquote)).where($\phi$(condition)). select(varList,\upquote count\upquote)
\\ 
            \bottomrule
        \end{tabular}
        }
        \end{small}
    \end{table}
    
\begin{table}[ht]
        \centering
    	\caption{Translation of Result Clause}
    	\label{tab:resultClauseTranslation}
    	\begin{small}
        	\resizebox{\textwidth}{!}{%
            \begin{tabular}{@{} L{4cm} @{~~~} L{8cm}}
            \toprule
            \textbf{Result Clause RC} & \textbf{Gremlin Traversal GT = $\phi$(RC)}\\ 
            \midrule
            SELECT varList QP         & $\phi$(QP).select(varList) \\ 
            \noalign{\smallskip}%
            ASK QP                    & $\phi$(QP).range(0,1).hasNext() \\ 
            \noalign{\smallskip}%
            DESCRIBE ?var QP          & $\phi$(QP).union(select(\upquote var\upquote).as(\upquote sub\upquote).$\phi$(var ?pred ?obj), select(\upquote var\upquote).as(\upquote obj\upquote).$\phi$(?sub ?pred var)).select(\upquote sub\upquote,\upquote pred\upquote,\upquote obj\upquote) \\ 
            \noalign{\smallskip}%
            CONSTRUCT \{?s ?p o\} QP  & $\phi$(QP).select(\upquote s\upquote).as(\upquote sub\upquote).select(\upquote p\upquote).as(\upquote pred \upquote).constant(\upquote o\upquote). as(\upquote obj\upquote).select(\upquote sub\upquote,\upquote pred\upquote,\upquote obj\upquote)\\ 
            \bottomrule
            \end{tabular}
            }
        \end{small}
    \end{table}

The proposed SPARQL to Gremlin translation module made a lot of improvements over the existing Gremlinator. The module could handle the standard SPARQL queries for RDF datasets along with variable predicates and unbalanced UNION queries. Along with these improvements, support for the following constructs is added:  
\begin{inparaenum}[i)]
    \item Property path expression: InversePath, SequencePath, AlternativePath, ZeroOrMorePath, OneOrMorePath, ZeroOrOnePath, NegatedPropertySet
    \item Functions on Numerics: ABS, ROUND, CEIL, FLOOR, RAND
    \item String functions: STRSTARTS, STRENDS, CONTAINS, REGEX, langMatches
    \item AGGREGATES: SUM, MIN, MAX, AVG, SAMPLE
    \item CONSTRUCT, DESCRIBE, ASK, BIND, VALUES.
\end{inparaenum}

We have made a certain improvement, but there are some limitations as well. Those are 
\begin{inparaenum}[i)]
    \item String functions: STRLEN, SUBSTR, UCASE, LCASE, STRBEFORE, STRAFTER, ENCODE\_FOR\_URI, CONCAT, REPLACE
    \item Dataset Definition: FROM, FROM NAMED and GRAPH
    \item Functions on Dates and Times- NOW, YEAR, MONTH, DAY, HOURS, MINUTES, SECONDS, TIMEZONE, TZ
    \item Hash Functions-MD5, SHA1, SHA256, SHA384, SHA512.
    \item SPARQL update: INSERT, DELETE.
\end{inparaenum} Most of these limitations are there because there are no equivalent gremlin operations for these clauses. 



%% file: files/Experiments.tex
\section{Experiments}
\label{sec:experiments}
\noindent
\textbf{Benchmark Datasets:} We used following three benchmark datasets for our evaluations.
\noindent
\textbf{Lehigh University Benchmark (LUBM):} LUBM~\cite{lubm} is synthetic OWL-DL dataset built around a university domain ontology and provides 14 benchmark queries. We generated data for 1000 universities which consist of around 130 million triples. 

\noindent
\textbf{Berlin SPARQL Benchmark (BSBM):} BSBM~\cite{bsbm} benchmark is built around e-commerce use case. We used the provided triple data generator to generate a dataset of 100M triples. The benchmark also provides query templates consisting and we generated 12 queries using the provided query templates containing  different constructs such as SELECT, CONSTRUCT, DESCRIBE, FILTER, REGEX, and BOUND. 

\noindent
\textbf{Waterloo SPARQL Diversity Test Suite (WatDiv):} WatDiv\cite{watdiv} is another Synthetic RDF Benchmark which consists of diversified stress testing workload. We used 100M triples WatDiv dataset for our experiments. It comes with 20 SPARQL Select query templates that can be divided into four categories: linear queries (L), star queries (S), snowflake-shaped queries (F) and complex queries (C). We generated one query from each of these templates for our experiments. 

\noindent
\textbf{Experimental Setup:} We implemented our proposed Data transformation and query translation modules on top of JanusGraph allowing us to store and query RDF on JanusGraph. We call this system ERGS.  We compared our system with Virtuoso\footnote{\url{https://virtuoso.openlinksw.com}}, BlazeGraph\footnote{\url{https://blazegraph.com}} and TinkerPop Sparql-Gremlin Plugin: Gremlinator. All the experiments were performed on RHEL machine with 128 GB RAM, 16 cores and 1TB HDD. For all the queries, the execution time out was set to 10 minutes. BlazeGraph and Virtuoso were run with all the recommended optimization settings and all the recommended indexes were built for both of them. We used JanusGraph as the Property Graph backend for Gremlinator. JanusGraph was run in the server mode using the TinkerPop gremlin server. For ERGS, Virtuoso and BlazeGraph standard RDF datasets and queries were used as provided by the datasets. For Gremlinator data ingested does not include the complete IRI for predicate and the SPARQL queries were manually translated to a format that can be consumed by Gremlinator.

\noindent
\textbf{Evaluation Metrics: } Following metrics were used for evaluating the performance of Query Engine. 
\textit{(i)} \textit{Cold Cache time}: it is the time taken for executing the query for the first time, when data is not already present in cache; 
 \textit{(ii) Warm Cache time}: it is the average time taken (averaged over 5 runs) for executing the query, when data is already present in cache; and  
\textit{(iii) Output Size}: it captures the number of rows returned as the result of query. 

\subsection{Results and Discussions}
We report the query runtimes achieved by different systems on the three different benchmarks considered. The aim of the experiments is to study the performance characteristics of the proposed system and see if Property Graphs can be used for storing and querying RDF data using our proposed transformations.

Table~\ref{tab:lubmResults} shows the query performance of different systems on LUBM benchmark. LUBM queries test the system for RDF retrieval as well as reasoning capabilities. We can see that ERGS produces complete set of results for all the queries except two queries that require reasoning beyond OWL2-QL and query runtimes are not far behind the native RDF stores. We do note that for three queries (Q6, Q9, Q14), our proposed system timed out and could not fetch the desired results. These three queries have a large output set (e.g. more than 7.9 million for Q14)  and this is the main reason for the slow execution time when compared to the native triple stores. This is becuase JanusGraph does not utilize \textit{locality of reference} and distributes the nodes across blocks, thus making the traversals for such a large number of nodes painfully slow. Other than these three queries, we note that ERGS was able to finish execution for most of the queries and is competent with the Triple stores. On the other hand, Gremlinator could not finish its execution for any of the queries. One of the main reasons behind this is Gremlinator's inability to use the underlying indexing strategies of Property Graphs.

\input{files/lubm-table.tex}

\input{files/bsbm-table.tex}

\input{files/watdiv-table.tex}

Table~\ref{tab:bsbmResults} summarizes the results for the benchmark queries on the BSBM dataset for all the four systems. We observe that ERGS could handle all the queries and is on par with the execution times and number of results with triples stores. Gremlinator, on the other hand, with its limitations could not handle many of the queries (denoted as unsupported) and timed out in cases where it was able to successfully translate the SPARQL queries. Likewise, for WatDiv dataset (Table~\ref{tab:watdivResults}), ERGS was able to handle all but one queries within the time limit, is competent and holds well against the native triple stores, and is orders of magnitude better than Gremlinator that could only handle a subset of the benchmark queries. 

These results indicate that ERGS, a Property Graph database to support RDF data using our proposed data transformation and query translation modules, was able to compete with specially designed RDF stores on most of the benchmark datasets. For many of the benchmark queries, the performance was very close to the RDF stores, and is orders of magnitude faster when compared with Gremlinator. Further, our proposed modules have a much broader coverage and can handle all the SPARQL constructs in an efficient manner when compared with Gremlinator.

%% file: files/lubm-table.tex
\begin{table}[t]
\centering
\begin{small}
\caption{LUBM Benchmark Results on 1000U data (exec. time in m.sec.)}
    \label{tab:lubmResults}
\resizebox{\textwidth}{!}{%
\begin{tabular}{@{}l rrr @{~~~} rrr @{~~~}crr @{~~~}rrr@{}}
\toprule
\multirow{2}{*}{\textbf{Queries}} & \multicolumn{3}{c}{\textbf{Virtuoso}} & \multicolumn{3}{c}{\textbf{Blazegraph}} & \multicolumn{3}{c}{\textbf{Gremlinator}} & \multicolumn{3}{c}{\textbf{ERGS}} \\
\cmidrule(r){2-4} \cmidrule(r){5-7} \cmidrule(r){8-10}\cmidrule(r){11-13} 
 & \textbf{\begin{tabular}[c]{@{}l@{}}Cold\\ Cache\end{tabular}} & \textbf{\begin{tabular}[c]{@{}l@{}}Warm\\ Cache\end{tabular}} & \textbf{\begin{tabular}[c]{@{}l@{}}o/p\\ Size\end{tabular}} & \textbf{\begin{tabular}[c]{@{}l@{}}Cold\\ Cache\end{tabular}} & \textbf{\begin{tabular}[c]{@{}l@{}}Warm\\ Cache\end{tabular}} & \textbf{\begin{tabular}[c]{@{}l@{}}o/p\\ Size\end{tabular}} & \multicolumn{1}{l}{\textbf{\begin{tabular}[c]{@{}l@{}}Cold\\ Cache\end{tabular}}} & \textbf{\begin{tabular}[c]{@{}l@{}}Warm\\ Cache\end{tabular}} & \textbf{\begin{tabular}[c]{@{}l@{}}o/p\\ Size\end{tabular}} & \textbf{\begin{tabular}[c]{@{}l@{}}Cold\\ Cache\end{tabular}} & \textbf{\begin{tabular}[c]{@{}l@{}}Warm\\ Cache\end{tabular}} & \textbf{\begin{tabular}[c]{@{}l@{}}o/p\\ Size\end{tabular}} \\ 
 \midrule
\textbf{Q1} & 593 & 2 & 4 & 475 & 14 & 4 & \multicolumn{3}{c}{Time Out} & 1079 & 32 & 4 \\
\textbf{Q2} & 11,249 & 8,245 & 2,528 & 122,068 & 118,372 & 2,528 & \multicolumn{3}{c}{Time Out} & \multicolumn{3}{c}{Time Out} \\
\textbf{Q3} & 748 & 4 & 6 & 7 & 5 & 6 & \multicolumn{3}{c}{Time Out} & 57 & 22 & 6 \\
\textbf{Q4} & 1,644 & 19 & 34 & 14 & 10 & 34 & \multicolumn{3}{c}{Time Out} & 829 & 159 & 34 \\
\textbf{Q5} & 209 & 205 & 146 & 16 & 11 & 719 & \multicolumn{3}{c}{Time Out} & 441 & 69 & 719 \\
\textbf{Q6} & 107,047 & 104,969 & 7,924,765 & 32,283 & 31,331 & 7,924,765 & \multicolumn{3}{c}{Time Out} & \multicolumn{3}{c}{Time Out} \\
\textbf{Q7} & 320 & 7 & 59 & 10 & 6 & 59 & \multicolumn{3}{c}{Time Out} & 108 & 31 & 59 \\
\textbf{Q8} & 390 & 380 & 5916 & 64 & 55 & 5,916 & \multicolumn{3}{c}{Time Out} & 6747 & 1853 & 5916 \\
\textbf{Q9} & 14,750 & 13,184 & 131,969 & 30,113 & 29,658 & 131,969 & \multicolumn{3}{c}{Time Out} & \multicolumn{3}{c}{Time Out} \\
\textbf{Q10} & 3 & 2 & 0 & 15 & 14 & 0 & \multicolumn{3}{c}{Time Out} & 15 & 12 & 0 \\
\textbf{Q11} & 45 & 7 & 224 & 8 & 6 & 224 & \multicolumn{3}{c}{Time Out} & 32 & 19 & 0 \\
\textbf{Q12} & 4 & 3 & 15 & 15 & 14 & 0 & \multicolumn{3}{c}{Time Out} & 58 & 55 & 0 \\
\textbf{Q13} & 365 & 185 & 4761 & 90 & 41 & 4,760 & \multicolumn{3}{c}{Time Out} & 7342 & 187 & 4760 \\
\textbf{Q14} & 103,336 & 103,654 & 7,924,765 & 31,135 & 30,759 & 7,924,765 & \multicolumn{3}{c}{Time Out} & \multicolumn{3}{c}{Time Out} \\
\bottomrule
\end{tabular}
}
\end{small}
\end{table}

%% file: files/bsbm-table.tex
\begin{table}[t]
\centering
\begin{small}
\caption{BSBM Benchmark Results (exec. time in m.sec.)}
    \label{tab:bsbmResults}
\resizebox{\textwidth}{!}{%
\begin{tabular}{@{}l rrr @{~~~} rrr @{~~~}crr @{~~~}rrr@{}}
\toprule
\multirow{2}{*}{\textbf{Queries}} & \multicolumn{3}{c}{\textbf{Virtuoso}} & \multicolumn{3}{c}{\textbf{Blazegraph}} & \multicolumn{3}{c}{\textbf{Gremlinator}} & \multicolumn{3}{c}{\textbf{ERGS}} \\
\cmidrule(r){2-4} \cmidrule(r){5-7} \cmidrule(r){8-10}\cmidrule(r){11-13} 
 & \textbf{\begin{tabular}[c]{@{}l@{}}Cold\\ Cache\end{tabular}} & \textbf{\begin{tabular}[c]{@{}l@{}}Warm\\ Cache\end{tabular}} & \textbf{\begin{tabular}[c]{@{}l@{}}O/P\\ Size\end{tabular}} & \textbf{\begin{tabular}[c]{@{}l@{}}Cold\\ Cache\end{tabular}} & \textbf{\begin{tabular}[c]{@{}l@{}}Warm\\ Cache\end{tabular}} & \textbf{\begin{tabular}[c]{@{}l@{}}O/P\\ Size\end{tabular}} & \multicolumn{1}{l}{\textbf{\begin{tabular}[c]{@{}l@{}}Cold\\ Cache\end{tabular}}} & \textbf{\begin{tabular}[c]{@{}l@{}}Warm\\ Cache\end{tabular}} & \textbf{\begin{tabular}[c]{@{}l@{}}O/P\\ Size\end{tabular}} & \textbf{\begin{tabular}[c]{@{}l@{}}Cold\\ Cache\end{tabular}} & \textbf{\begin{tabular}[c]{@{}l@{}}Warm\\ Cache\end{tabular}} & \textbf{\begin{tabular}[c]{@{}l@{}}O/P\\ Size\end{tabular}} \\ 
 \midrule
\textbf{Q1} & 528 & 4 & 0 & 468 & 38 & 0 & \multicolumn{3}{c}{Time Out} & 655 & 73 & 0 \\
\textbf{Q2} & 927 & 7 & 18 & 51 & 22 & 18 & \multicolumn{3}{c}{Unsupported} & 105 & 71 & 18 \\
\textbf{Q3} & 67 & 5 & 0 & 52 & 27 & 0 & \multicolumn{3}{c}{Unsupported} & 48 & 49 & 0 \\
\textbf{Q4} & 145 & 18 & 0 & 74 & 34 & 0 & \multicolumn{3}{c}{Unsupported} & 121 & 110 & 0 \\
\textbf{Q5} & 47 & 33 & 5 & 490 & 353 & 5 & \multicolumn{3}{c}{Unsupported} & 3,620 & 3,211 & 5 \\
\textbf{Q6} & 1,376 & 814 & 9 & 2,807 & 2,436 & 9 & \multicolumn{3}{c}{Unsupported} & \multicolumn{1}{c}{1,244} & 1220 & 9 \\
\textbf{Q7} & 4,772 & 10 & 2 & 20 & 16 & 2 & \multicolumn{3}{c}{Unsupported} & 78 & 76 & 2 \\
\textbf{Q8} & 1,104 & 6 & 0 & 29 & 28 & 0 & \multicolumn{3}{c}{Time Out} & 34 & 31 & 0 \\
\textbf{Q9} & 7 & 2 & 24 & 16 & 5 & 24 & \multicolumn{3}{c}{Time Out} & 38 & 20 & 24 \\
\textbf{Q10} & 4,218 & 6 & 0 & 32 & 26 & 0 & \multicolumn{3}{c}{Time Out} & 289 & 285 & 0 \\
\textbf{Q11} & 9 & 1 & 10 & 14 & 3 & 10 & \multicolumn{3}{c}{Unsupported} & 19 & 19 & 10 \\
\textbf{Q12} & 4,385 & 6 & 8 & 15 & 7 & 8 & \multicolumn{3}{c}{Time Out} & 60 & 53 & 8 \\
\bottomrule
\end{tabular}
}
\end{small}
\end{table}

%% file: files/watdiv-table.tex
\begin{table}[t]
\centering
\begin{small}
\caption{Watdiv Benchmark Results (exec. time in m.sec.)}
    \label{tab:watdivResults}
\resizebox{\textwidth}{!}{%
\begin{tabular}{@{}l rrr @{~~~} rrr @{~~~}crr @{~~~}rrr@{}}
\toprule
\multirow{2}{*}{\textbf{Queries}} & \multicolumn{3}{c}{\textbf{Virtuoso}} & \multicolumn{3}{c}{\textbf{Blazegraph}} & \multicolumn{3}{c}{\textbf{Gremlinator}} & \multicolumn{3}{c}{\textbf{ERGS}} \\
\cmidrule(r){2-4} \cmidrule(r){5-7} \cmidrule(r){8-10}\cmidrule(r){11-13} 
 & \textbf{\begin{tabular}[c]{@{}l@{}}Cold\\ Cache\end{tabular}} & \textbf{\begin{tabular}[c]{@{}l@{}}Warm\\ Cache\end{tabular}} & \textbf{\begin{tabular}[c]{@{}l@{}}o/p\\ Size\end{tabular}} & \textbf{\begin{tabular}[c]{@{}l@{}}Cold\\ Cache\end{tabular}} & \textbf{\begin{tabular}[c]{@{}l@{}}Warm\\ Cache\end{tabular}} & \textbf{\begin{tabular}[c]{@{}l@{}}o/p\\ Size\end{tabular}} & \multicolumn{1}{l}{\textbf{\begin{tabular}[c]{@{}l@{}}Cold\\ Cache\end{tabular}}} & \textbf{\begin{tabular}[c]{@{}l@{}}Warm\\ Cache\end{tabular}} & \textbf{\begin{tabular}[c]{@{}l@{}}o/p\\ Size\end{tabular}} & \textbf{\begin{tabular}[c]{@{}l@{}}Cold\\ Cache\end{tabular}} & \textbf{\begin{tabular}[c]{@{}l@{}}Warm\\ Cache\end{tabular}} & \textbf{\begin{tabular}[c]{@{}l@{}}o/p\\ Size\end{tabular}} \\ 
 \midrule
\textbf{L1} & 311 & 2 & 6 & 384 & 14 & 6 & \multicolumn{3}{c}{Time Out} & 378 & 25 & 6 \\
\textbf{L2} & 106 & 5 & 106 & 95 & 31 & 106 & \multicolumn{3}{c}{Unsupported} & \multicolumn{1}{c}{417} & 397 & 106 \\
\textbf{L3} & 2 & 2 & 27 & 11 & 8 & 27 & \multicolumn{3}{c}{Time Out} & 21 & 12 & 27 \\
\textbf{L4} & 18 & 15 & 606 & 74 & 36 & 606 & 372115 & 317436 & 606 & 480 & 466 & 606 \\
\textbf{L5} & 46 & 43 & 1310 & 249 & 87 & 1310 & \multicolumn{3}{c}{Unsupported} & 1723 & 1590 & 1310 \\
\midrule
\textbf{S1} & 709 & 224 & 4 & 15 & 17 & 4 & 395651 & 393990 & 4 & \multicolumn{1}{c}{50} & 48 & 4 \\
\textbf{S2} & 8 & 8 & 114 & 49 & 24 & 114 & \multicolumn{3}{c}{Time Out} & 1084 & 1076 & 114 \\
\textbf{S3} & 1543 & 1506 & 3189 & 156 & 106 & 3189 & 338252 & 336151 & 3189 & 2350 & 2338 & 3189 \\
\textbf{S4} & 282 & 5 & 8 & 65 & 38 & 8 & \multicolumn{3}{c}{Time Out} & \multicolumn{1}{c}{5045} & 4817 & 8 \\
\textbf{S5} & 6 & 5 & 0 & 56 & 30 & 0 & \multicolumn{3}{c}{Time Out} & 1074 & 1071 & 0 \\
\textbf{S6} & 20 & 5 & 21 & 22 & 12 & 21 & \multicolumn{3}{c}{Time Out} & 88 & 85 & 21 \\
\textbf{S7} & 1 & 1 & 1 & 4 & 3 & 1 & 338819 & 338030 & 1 & 10 & 9 & 1 \\

\midrule
\textbf{F1} & 490 & 7 & 0 & 45 & 43 & 0 & \multicolumn{3}{c}{Time Out} & 572 & 530 & 0 \\
\textbf{F2} & 541 & 85 & 139 & 62 & 63 & 139 & \multicolumn{1}{l}{397849} & 378560 & 139 & 1277 & 1286 & 139 \\
\textbf{F3} & 213 & 30 & 211 & 48 & 33 & 211 & \multicolumn{3}{c}{Time Out} & 553 & 537 & 211 \\
\textbf{F4} & 144 & 138 & 234 & 48 & 30 & 234 & \multicolumn{3}{c}{Unsupported} & 1166 & 1155 & 234 \\
\textbf{F5} & 401 & 7 & 43 & 17 & 10 & 43 & \multicolumn{1}{l}{373072} & 370873 & 43 & 39 & 34 & 43 \\
\midrule
\textbf{C1} & 1765 & 189 & 201 & 473 & 161 & 201 & \multicolumn{3}{c}{Unsupported} & 26522 & 26472 & 201 \\
\textbf{C2} & 2232 & 446 & 22 & 1128 & 840 & 22 & \multicolumn{3}{c}{Unsupported} & 86642 & 84548 & 22 \\
\textbf{C3} & 42375 & 39612 & 4244261 & 7031 & 6980 & 4244261 & \multicolumn{3}{c}{Time Out} & \multicolumn{3}{c}{Time Out} \\
\bottomrule
\end{tabular}
}
\end{small}
\end{table}

%% file: files/conclusions.tex
\section{Conclusions}
\label{sec:conclusions}
We described data transformation and query translation processes that can allow a Property Graph database to store RDF data and offer SPARQL querying in an efficient manner. Empirical evaluation using three different benchmarks confirmed the superiority of our propsed solution over Gremlinator, the current state-of-the-art for supporting RDF data on Property Graphs. We also found the query runtime performance of the proposed system to be competent with native triple stores (Virtuoso and Blazegraph). Thus, we beleive that our proposed solution offers a reasonable first solution to supporting RDF datasets on Property Graph databases. We plan to continue to further improve the performance by better indexing and cache management to handle large node scan queries which is one of the main bottlenecks observed during our benchmarking experiments. Further, since currently we only support limited reasoning using query expansion, we plan to add forward-chaining support and support for more expressive reasoning to potentially handle more expressive OWL profiles.